%% file: main.tex
\documentclass[a4paper]{article}
\usepackage[preprint]{neurips_2024}

\usepackage{graphicx}
\usepackage{times}
\usepackage{latexsym}
\usepackage[hyphens]{url} 
\usepackage{hyperref}
\usepackage{caption} 
\usepackage{array} 
\usepackage{booktabs}
\usepackage{float}
\usepackage{comment}
\usepackage{amsmath,amsthm,amsfonts,amssymb,bm,stmaryrd}
\usepackage{makecell} 
\usepackage{multirow, multicol}
\usepackage{colortbl}
\usepackage{xcolor}
\usepackage{wrapfig} 
\usepackage{pdflscape}
\usepackage[most]{tcolorbox}
\usepackage{makecell}

\title{Motif 2.6B Technical Report}
\author{Motif Technologies}

\begin{document}

\maketitle

\begin{abstract}
Recent advancements in Large Language Models (LLMs) have revolutionized artificial intelligence, yet developing an effective foundational LLM that balances high performance with computational efficiency remains challenging, especially for emerging research groups. To address this gap, we introduce Motif-2.6B, a 2.6-billion-parameter foundation model designed to democratize advanced LLM capabilities. Motif-2.6B incorporates several innovative architectural enhancements, including Differential Attention and PolyNorm activation functions, which improve long-context comprehension, reduce hallucination, and enhance in-context learning capabilities. We rigorously tested multiple novel architectural components through extensive experimentation to determine the optimal architecture for Motif-2.6B.
Comprehensive evaluations demonstrate that Motif-2.6B consistently meets or exceeds the performance of similarly sized state-of-the-art models across diverse benchmarks, showcasing its effectiveness, scalability, and real-world applicability. Through detailed experiments and tailored techniques, Motif-2.6B significantly advances the landscape of efficient, scalable, and powerful foundational LLMs, offering valuable insights and a robust foundation for future research and deployment.
\end{abstract}

\input{introduction}

\input{architecture}

\input{pretraining}

\input{posttraining}

\bibliographystyle{plain}
\bibliography{reference}

\clearpage
\input{appendix}

\end{document}

%% file: introduction.tex
\section{Introduction}
Recent advances in Large Language Models (LLMs) have significantly reshaped the landscape of artificial intelligence, powering a wide array of applications, from ChatBots and automated assistants to sophisticated reasoning engines and content generators, demonstrating their versatility and impact in the real-world~\cite{dam2024complete, raza2025industrial}. However, despite this growth of powerful models, developing a foundational LLM from scratch, particularly one with manageable parameter size that retains robust capabilities, continues to pose a major challenge for emerging research groups. Our work seeks to fill this gap by introducing \textbf{Motif-2.6B}, a 2.6‑billion‑parameter foundation model that aims to deliver competitive performance while remaining efficient and scalable.

In this work, we present Motif-2.6B as an essential step forward in the democratization of powerful LLM technologies, balancing computational affordability with functional versatility. With 2.6 billion parameters, 
Motif-2.6B bridges the gap between smaller-scale models and the mega-models typically requiring substantial infrastructure investments. This report provides comprehensive insight into the architecture choices, training methodologies, data curation processes, and performance benchmarks of Motif-2.6B, contributing valuable knowledge for future research and development in efficient language modeling.

Distinctively, Motif-2.6B is characterized by extensive experimentation and integration of novel architectural components. Our exploration includes novel techniques such as QK normalization \cite{henry2020query}, differential attention \cite{ye2024differential}, cross-layer attention \cite{mu2024cross}, normalized GPT \cite{loshchilov2024ngpt}, and polynomial activation \cite{zhuo2024polynomial}. Through rigorous testing, we have selectively incorporated some of these advanced structures, enabling Motif-2.6B to achieve unique performance characteristics and enhanced efficiency. These improvements position Motif-2.6B not only as a robust foundation for various downstream tasks but also as a significant advancement in the ongoing evolution of LLM architecture design.

Our pre-training process utilized approximately 2.5 trillion tokens of training data. Following recent trends in LLM pre-training \cite{deng2025plm, yang2025qwen3, granite2024granite, grattafiori2024llama, allal2025smollm2}, we initially emphasized general English content and progressively increased the proportion of domain-specific data such as coding, mathematics, and academic literature towards the later training stages. While conventional approaches often introduce discrete, abrupt changes in data composition, our approach employed a linear, gradual scheduling method for data mixture, smoothly transitioning the proportions of various data types throughout the entire pre-training phase. This novel scheduling strategy enabled Motif-2.6B to steadily adapt to complex linguistic and domain-specific knowledge without sudden disruptions.

% 위 Following recent trends in LLM pre-training 에 해당하는 citation 찾아야함. 일단 Qwen3.

In the post-training, we focused on enhancing model performance via systematic dataset refinement and effective optimization strategies. We generated high-quality synthetic datasets through rejection sampling and effectively combined multiple distinct datasets through a fusion approach. We also conducted quality filtering to ensure the integrity and relevance of training data. Furthermore, extensive exploration and iterative experimentation allowed us to identify and curate an optimal collection of publicly available datasets. Finally, we employed Direct Preference Optimization (DPO) \cite{rafailov2023direct} to align Motif-2.6B closely with practical human preferences and downstream task requirements, significantly boosting the model's generalization and effectiveness in various real-world applications.

For evaluation, we rigorously assessed Motif-2.6B against various competitive benchmarks. Recognizing that existing models often report benchmarks with varying settings, such as different numbers of shots (n-shot) or inclusion of Chain-of-Thought (CoT) reasoning, we carefully replicated each model's reported evaluation settings to ensure a fair and consistent comparison. Through this carefully designed approach, we validated that Motif-2.6B consistently matches or surpasses the performance of similarly sized, globally leading models, confirming its effectiveness and robustness across diverse tasks.

In the following sections, we first provide a detailed description of the architectural design of Motif-2.6B, including the novel components incorporated into the model. Next, we outline the specific methodologies employed during the pre-training and post-training phases. We then present comprehensive evaluations of the model's performance across various benchmarks. Finally, we conclude by summarizing our key insights and discussing potential avenues for future research and development.

%% file: architecture.tex
\section{Architecture}
\begin{figure*}[h!]
    \centering
    \includegraphics[width=1\linewidth]{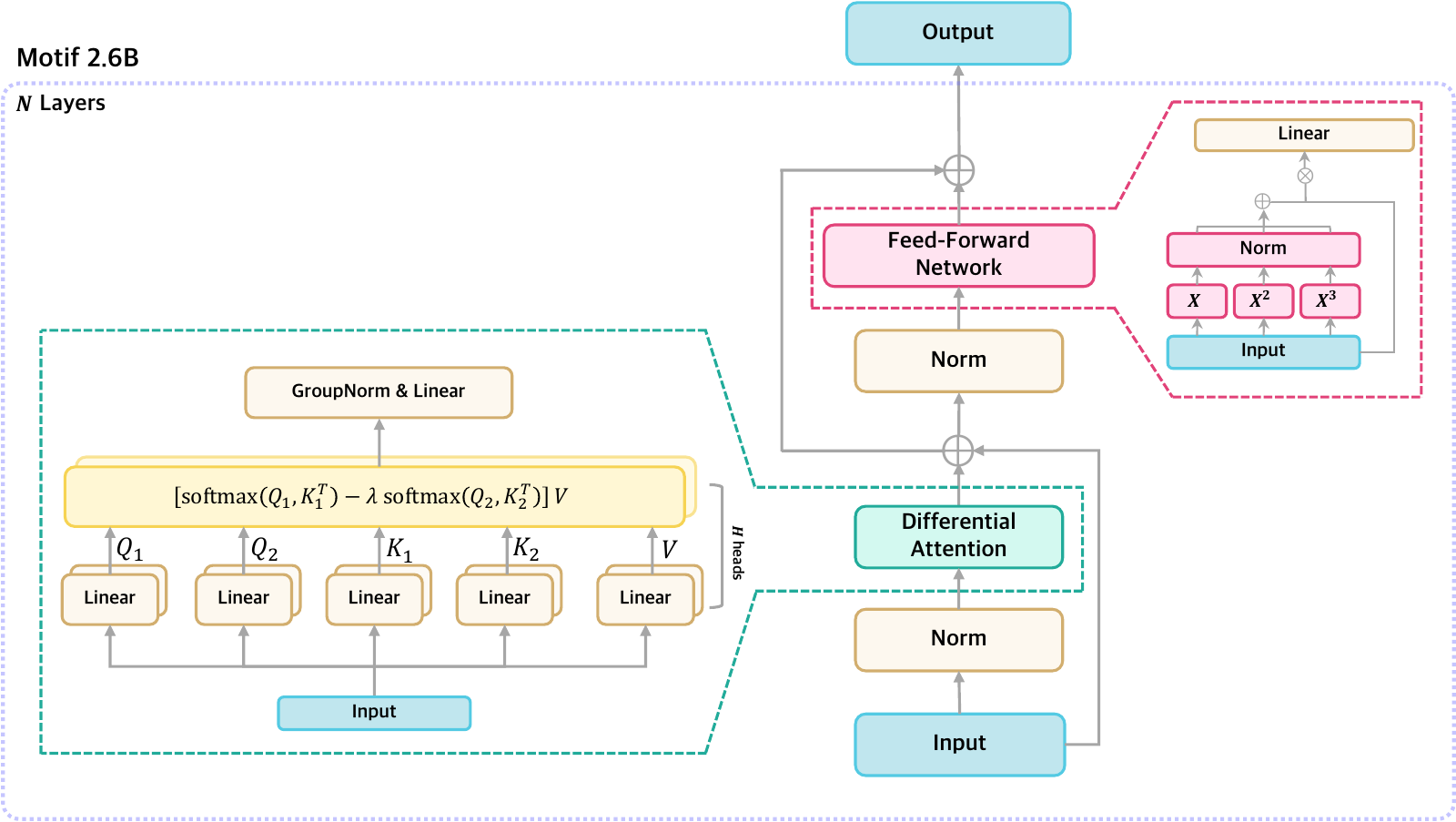}
    \caption{An illustration of the architecture of Motif-2.6B. Motif-2.6B employs Differential Attention and PolyNorm layers in place of conventional attention and normalization mechanisms. These components were selected based on a systematic evaluation of individual architectural variants, with the goal of enhancing representation and improving model robustness.}
    \label{fig:model_archtecture}
\end{figure*}

\begin{wrapfigure}{r}{0.52\textwidth} 
    \centering
    \vspace{-17pt}
    \begin{tabular}{l|c}
        \toprule
        \textbf{Model size} & \textbf{2.6B} \\
        \midrule
        Hidden dimension & 2048 \\
        Number of layers & 32 \\
        Feed-Forward dimension & 8192 \\
        Number of heads & 16 \\
        Number of KV heads & 16 \\
        \midrule
        Attention Mechanism & Differential Attention \\
        Activation Function & PolyNorm \\
        Max Sequence Length & 4096, 16384 \\
        Positional Embeddings & RoPE ($\theta=500,000$) \\
        Vocab size & 219,520 \\
        Tied word embedding & True \\
        \bottomrule
    \end{tabular}
    \captionof{table}{The detailed hyper-parameters of Motif-2.6B. The standard configuration supports a sequence length of 4,096 tokens, while Motif-2.6B-LC extends this capacity to 16,384 tokens to enable long-context processing. Here, ``LC'' denotes ``Long Context''.}
    \label{tab:hyper-params} 
\end{wrapfigure}

\subsection{Design Decision for Final Model Architecture}

Motif-2.6B is based on a decoder-only Transformer architecture consisting of approximately 2.6 billion parameters, with around 2.2 billion dedicated to non-embedding components. Although fundamentally adhering to the traditional decoder-only Transformer structure, the final architecture of Motif-2.6B was carefully determined through an extensive experimentation process, designed to explore and validate numerous recent architectural innovations. Specifically, we systematically evaluated each architectural variant individually, including QK normalization \cite{henry2020query}, differential attention \cite{ye2024differential}, cross-layer attention \cite{mu2024cross}, normalized GPT \cite{loshchilov2024ngpt}, and polynomial activation \cite{zhuo2024polynomial}. Following these individual assessments, we conducted additional combinational experiments to investigate potential synergistic effects between selected methods. Each experimental configuration was carefully tested across multiple benchmarks, using diverse model sizes (0.6B, 1.8B, and 4.6B parameters) and various dataset scales under a strictly controlled computational budget of $3 \times 10^{20}$ FLOPs. Based on comprehensive benchmarking results, the most effective combination emerged clearly, leading us to select an architecture that integrates Differential Attention with polynomial activation, referred to as PolyNorm as illustrated in Figure~\ref{fig:model_archtecture}.

\textbf{PolyNorm} is a variant of polynomial composition activation functions designed to allow the model to capture more fine-grained and higher-order relationships between tokens. we constrained the maximum degree of the composed polynomial functions to $3$.

\textbf{Differential Attention} enhances focus on the relevant context by computing two distinct attention maps and subtracting one from the other, thus canceling noise. This mechanism produces more spare and precise attention patterns. Empirical results show substantial improvements in long-context comprehension, information retrieval, hallucination mitigation, and in-context learning.

This combination leads to improvements in model performance, training stability, and generalization, while requiring fewer parameter and tokens. The detailed hyper-parameters of Motif-2.6B is described in Table~\ref{tab:hyper-params}.

As existing frameworks currently lack optimized support for our model architecture, we trained our models using an internally developed framework featuring custom HIP-based kernels optimized for the ROCm platform on AMD GPUs. Given that our proprietary in-house framework is not publicly available, we are concurrently developing alternative kernels based on \href{https://github.com/huggingface/kernels}{HuggingFace kernels} package to enhance accessibility and broader adoption. These kernels can be downloaded and utilized from the links: \href{https://huggingface.co/Motif-Technologies/activation}{https://huggingface.co/Motif-Technologies/activation} and \href{https://huggingface.co/Motif-Technologies/optimizer}{https://huggingface.co/Motif-Technologies/optimizer}.

\subsection{Tokenizer}

We begin with the \texttt{$\text{o200k}\_\text{base}$} tokenizer from tiktoken\footnote{\url{https://github.com/openai/tiktoken}}, which provides broad multilingual coverage. To enhance alignment with linguistic characteristics of Korean, we additionally trained a tokenizer on a curated in-house Korean corpus, thereby enabling the extraction of vocabulary items more closely aligned with Korean morphology and syntax.

The training yielded approximately 19,000 Korean-specific tokens, which were subsequently merged with the original \texttt{$\text{o200k}\_\text{base}$} vocabulary to form a unified tokenizer. The final tokenizer comprises 219,520 tokens, including special tokens, and improves the bytes-per-token for Korean by $12.6\%$, without compromising performance in English and other languages.

%% file: pretraining.tex
\section{Pre-training}

\subsection{Pre-training Dataset}

We constructed a high-quality pre-training corpus by aggregating diverse and rigorously filtered open-source pre-training datasets, including DCLM \cite{li2024datacomp}, TxT360 \cite{tang2024txt360}, Fineweb2 \cite{penedo2025fineweb2pipelinescale}, and FineMath \cite{allal2025smollm2}, covering a wide range of domains and modalities. In addition, given the lack of high-quality Korean-language datasets, we collected and curated an in-house Korean corpus to ensure better linguistic coverage and fidelity.

\subsection{Two-stage Training}

\subsubsection{Stage 1: Dynamic Mixture Pre-training}

\begin{table}[ht!]
    \centering
    \resizebox{\linewidth}{!}{
    \begin{tabular}{l|ccc|cc|ccc|c}
\toprule
& \multicolumn{3}{c|}{\textbf{Language Data}} & \multicolumn{2}{c|}{\textbf{Domain-specific}} & \multicolumn{3}{c|}{\textbf{Reasoning-Oriented}} & \multirow{2}{*}{Total} \\
& General Web & Multilingual & Korean & Academic & Specialized & Code & Math & Reasoning & \\
\midrule
Initial mixture ratio & 0.68 & 0.07 & 0.01 & 0.06 & 0.05 & 0.10 & 0.02 & 0.01 & 1 \\
Final mixture ratio   & 0.33 & 0.00 & 0.30 & 0.03 & 0.01 & 0.18 & 0.10 & 0.05 & 1 \\
\bottomrule
\end{tabular}
}
\caption{A comparison of initial and final dataset mixture ratios across categories during Stage 1. The final mixture reflects increased emphasis on Korean, code, and math data, while reducing the proportion of general web and multilingual content.}
\label{tab:dataset_ratios}

\end{table}

\paragraph{Dataset Mixture Strategy.} Pre-training was conducted in two stages: Stage 1 primarily focused on English web data to build strong generalization capabilities, while annealing stage shifted toward more specialized domains such as mathematics and code, targeting higher-level reasoning and problem-solving skills. The shift in dataset composition between stages required dynamic adaptation during training as summarized in Table~\ref{tab:dataset_ratios}.

To address this, we introduced a data mixing scheduler, conceptually analogous to a learning rate scheduler. Just as a learning rate scheduler adjusts optimization dynamics over time, our data mixing scheduler adjusts the domain composition during pre-training. Specifically, our training dataset was partitioned into eight domain groups, and their sampling ratios were adjusted linearly\textemdash either increasing or decreasing\textemdash to guide the model's exposure across training iterations.

Through experiments, we observed that a linear data mixing schedule consistently outperformed a fixed mixture ratio setting. In particular, models trained with dynamic scheduling achieved higher average benchmark scores, indicating that gradual adaptation to the evolving dataset composition enhances pre-training efficiency and final performance.

\subsubsection{Stage 2: Annealing Phase}

\paragraph{RoPE-Based Context Extension.} We extended the context length capacity for the Motif-2.6B-LC model by increasing the RoPE base frequency from 10,000 to 500,000 using the Adjusted Base Frequency (ABF) technique \cite{xiong2023effectivelongcontextscalingfoundation}, and correspondingly increased the maximum sequence length from 4,096 to 16,384 tokens. This long-context extension was applied during the final 80 billion tokens of the overall pre-training, enabling effective training on significantly longer sequences. The standard Motif-2.6B model was not subjected to long-context training and retains the original maximum sequence length of 4,096 tokens.

\paragraph{Simple Moving Average.} Additionally, to enhance training stability and convergence, we adopted a Simple Moving Average~(SMA) strategy for model weights \cite{li2025modelmergingpretraininglarge}. Concretely, every 8B training tokens, we collected the weights from the six most recent checkpoints and computed their element-wise average. The averaged model was directly fed back into the training loop without interruption. By smoothing over recent model snapshots in this way, the mechanism helps reduce training noise and mitigate abrupt parameter shifts, contributing to more stable optimization in the late stage of training.

In Stage 1, we trained approximately 2T tokens, focusing on broad domain coverage to establish strong generalization capabilities. During the annealing phase, training continued on an additional 25\% of the Stage 1 token volume, selectively emphasizing higher-quality and domain-specific data. We use the AdamW optimizer \cite{kingma2017adammethodstochasticoptimization} with a learning rate of \( 5 \times 10^{-4} \), \((\beta_1, \beta_2) = (0.9, 0.95)\), a weight decay of 0.1, and 5,000 warmup steps. Initial RoPE base frequency \(\theta\) is set to $10,000$. Pre-training is performed with a batch size of 4M tokens.

For learning rate scheduling, we adopted the Warmup-Stable-Decay~(WSD)~\cite{hu2024minicpmunveilingpotentialsmall}, in which the learning rate remains at its peak for the initial training phase and then decays proportionally to the inverse square root of the training step. Specifically, the learning rate is held at for the first 2T tokens and gradually decreases to (25\% of the peak) over the final 0.5T tokens, corresponding to the annealing phase.

%% file: posttraining.tex
\section{Post-training}

\begin{figure*}[hbt!]
    \centering
    \includegraphics[width=0.9\linewidth]{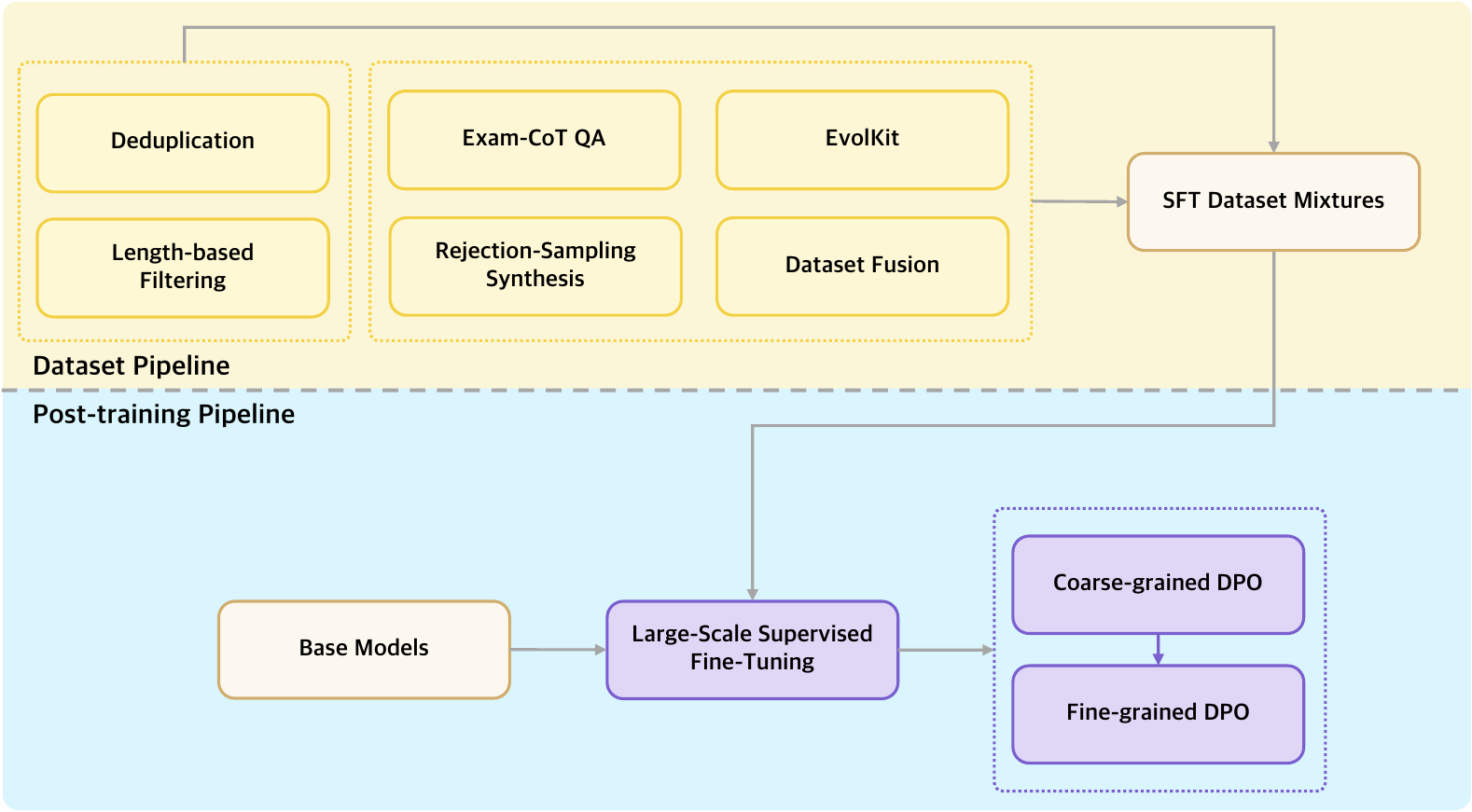}
    \caption{An illustration of the dataset and post-training pipeline, where carefully generated synthetic data and pre-processed existing datasets are merged into SFT dataset mixtures for large-scale supervised fine-tuning, followed by a two-stage alignment comprising coarse-grained and fine-grained phases.}
    
    % An illustration of dataset and post-training pipeline. SFT dataset mixtures are combined with carefully generated synthetic data and pre-processed existing datasets and used for large‑scale supervised fine‑tuning, which is followed by a two‑stage alignment comprising coarse‑grained and fine‑grained phases.}
    \label{fig:enter-label}
\end{figure*}

\subsection{Pre-processing}

\begin{table}[h!]
    \centering
    \resizebox{\textwidth}{!}{%
    \begin{tabular}{lcccccccc}
        \toprule
        \multirow{2}{*}{\textbf{Sequence Length}} & \multirow{2}{*}{\textbf{Average}} & \multicolumn{2}{c}{\textbf{Knowledge}} & \multicolumn{2}{c}{\textbf{Reasoning}} & \textbf{Instruction} & \multicolumn{2}{c}{\textbf{Math}} \\
        \cmidrule(lr){3-4} \cmidrule(lr){5-6} \cmidrule(lr){7-7} \cmidrule(lr){8-9}
        & & MMLU (MC) & MMLU Pro & BBH & ARC-C & IFEval & Math-Hard & Math500 \\
        \midrule
        0--4096 & 30.47 & 43.81 & 17.97 & \textbf{34.75} & \textbf{36.86} & \textbf{59.60} & \textbf{4.53} & 15.80 \\
        256--4096 & \textbf{30.88} & \textbf{44.30} & \textbf{18.68} & 33.34 & 35.58 & 59.49 & 4.38 & \textbf{20.40} \\
        1024--4096 & 24.73 & 41.75 & 16.32 & 32.32 & 33.45 & 50.96 & 3.10 & 9.20 \\
        \bottomrule
    \end{tabular}%
    }
    \caption{The impact of sequence length filtering on benchmark performance. Filtering out short sequences (e.g., $< 1024$ tokens) significantly reduces dataset size but degrades performance, particularly in reasoning and math tasks.}
    \label{tab:LengthFiltering}
\end{table}

High-quality data is a critical component of dataset pre-processing. We therefore applied two standard strategies\textemdash\textit{deduplication} and \textit{length filtering}.

\textbf{Length Filtering.} As introduced in \cite{shen2024rethinking}, emulating human-like conversational structure is especially important for SFT. This suggests shifting the emphasis from further ``knowledge acquisition''\textemdash already largely addressed during pre-training\textemdash to improving interaction quality and natural dialogue flow. To quantify this effect, we evaluated different lower bounds on sequence length. As depicted in Table~\ref{tab:LengthFiltering}, Removing only short sequences (e.g., $<256$ tokens) slightly improved the overall average score (30.47 $\rightarrow$ 30.88) and Math500 (15.80 $\rightarrow$ 20.40), whereas an aggressive cut-off at $<1024$ tokens substantially degraded performance across categories (e.g., Average $30.88 \rightarrow 24.73$). Based on this observation, we adopted a minimal threshold and discarded only trivially very short sequences ($<24$ tokens), which were unlikely to constitute complete or contextually meaningful interactions.

\textbf{Deduplication.} We employed a two-stage procedure. Exact duplicates were eliminated via straightforward string matching. Near-duplicates were detected using MinHash-based locality-sensitive hashing with Jaccard similarity $\ge 0.9$. Applied to roughly 4 million samples, this process removed about 200,000 redundant examples, improving dataset quality without significantly reducing its size.

\subsection{Post-training Dataset}

One of the most challenging tasks of post-training is the construction of a training dataset. Numerous strategies for constructing a high-quality training dataset remain one of the most challenging tasks in the post-training phase. Building upon proven methods from recent high-quality open-source initiatives \cite{lambert2024t}, our dataset strategy emphasized diversity across multiple dimensions, including domain coverage and task difficulty. To achieve this, we employed a balanced mixture of human-annotated data and rigorously filtered synthetic data generated by advanced LLMs.

Instead of assembling our dataset from scratch, we strategically leveraged existing curated mixtures from reputable sources such as Tulu \cite{lambert2024tulu3} and Granite \cite{granite2024granite}. These external datasets applied additional stages of careful filtering and refinement to ensure balanced representation across domains, optimal data quality, and seamless compatibility with our pre-training corpus.

\paragraph{SFT Dataset Mixture.} The SFT dataset were designed to enhance the model's instruction-following capabilities across multiple domains, including general language understanding, mathematical reasoning, and code generation. We paid special attention to domain balancing, drawing on domain ratio heuristics from reference models \cite{granite2024granite, yang2025qwen3, grattafiori2024llama}, while keeping the dataset within manageable size. To this end, the total number of tokens were limited to under 15B, targeting approximately 5M samples. Although we decided to use as much high-quality data as possible, the final size was chosen based on practical training time considerations as well as reference training volumes as in \cite{lambert2024tulu3, grattafiori2024llama}.

\paragraph{Alignment Dataset Mixture.} The alignment dataset was curated to align the model with human preferences, with a focus not just on helpfulness, harmlessness, and safety but also to ensure diversity. We employed distinct datasets for each stage to apply two-stage training strategy.

In Stage 1, we adopted the alignment datasets as introduced in Tulu~3~\cite{lambert2024tulu3}. Specifically, we combined two major subsets\textemdash the \texttt{Tulu 3 8B Preference Mixture}\footnote{\url{https://huggingface.co/datasets/allenai/llama-3.1-tulu-3-8b-preference-mixture}} and \texttt{Tulu 3 405B Preference Mixture}\footnote{\url{https://huggingface.co/datasets/allenai/llama-3.1-tulu-3-405b-preference-mixture}} and applied deduplication and remove harmful content to improve data quality and minimize redundancy.

In Stage 2, we constructed a new dataset by merging examples from \texttt{MagpieLM}\footnote{\url{https://huggingface.co/datasets/Magpie-Align/MagpieLM-DPO-Data-v0.1}} and the \texttt{LM-Sys}\footnote{\url{https://huggingface.co/datasets/mlabonne/lmsys-arena-human-preference-55k}}, then carefully removed harmful contents.

This shift from broadly representative preferences in Stage 1 to more curated, high-quality examples in Stage 2 illustrates our approach of establishing general alignment capabilities before training with precise preference signals.

\subsection{Synthetic Data}
We generated synthetic data by using proprietary models and open-source models. We analyzed the weakness of the model through evaluation and generated synthetic data that could address this weakness and enrich existing data with more comprehensive information. Exam-CoT QA dataset and prompt-evolution-based dataset are synthesized. Moreover, we experimented with a method which combines multiple data samples into one, namely ``dataset fusion''.
\paragraph{Exam-CoT QA.} We constructed a large-scale synthetic dataset comprising approximately 5 million multiple-choice question samples in standardized exam format. These samples were generated from diverse web-based sources spanning domains such as history, law, medicine, literature, religion, and other STEM-related fields. To enhance the model’s reasoning capability, each question was paired with a step-by-step rationale in the CoT style, guiding the model through intermediate reasoning before arriving at the final answer. This approach ensures that the dataset not only tests factual recall but also promotes structured reasoning.
\paragraph{EvolKit.}\footnote{\url{https://github.com/arcee-ai/EvolKit}} While the SFT dataset mixture was carefully designed to enhance various capabilities of the model, it might still have suffered from limitations such as simplicity, low diversity, and a large portion of examples of low difficulty. These constraints might limit the model’s capacity for deep reasoning and complex problem-solving. To this end, we enriched the existing SFT dataset mixture by incorporating additional instruction data synthesized by EvolKit, which leverages the Auto Evol-Instruct~\cite{zeng2024automatic} framework and employs \textit{Qwen3-8B} as the inference model.
\paragraph{Dataset Fusion.} We from the idea of teaching various types of knowledge to the model in a more compressed way by combining multiple samples into one. However, we found that the model generating fusion data needs capabilities such as multi-perspective thinking, multi-domain understanding, reasoning, and creativity. And these qualities are assumed to exist internally within larger LLMs but often lacks in smaller ones. \textit{Qwen3-8B} was initially employed to generate fusion texts, it tended merely to concatenate the given inputs. Therefore, we had to use \textit{GPT-4o} to generate fusion data aligned with our original intention. Rather than simply generating text from large LLMs for domains where data is scarce, we believe that distilling capabilities exclusively available in these models is a more effective approach to constructing high-quality synthetic datasets. The prompt for the dataset fusion is carefully designed to instruct a model to creatively fuse multiple conversational datasets into a single, cohesive message sequence. Rather than simple concatenation, the AI is guided to deeply understand each dataset, identify hidden thematic or conceptual connections, and integrate them into imaginative new scenarios that naturally blend their original intents. The resulting dataset emphasizes organic coherence, subtlety, and novel relational insights across blended domains. 
\paragraph{Rejection-Sampling Synthesis.} We constructed high-quality synthetic datasets using a rejection-sampling pipeline. For each prompt, an LLM generated multiple candidate instances, which were scored by a reward model; candidates whose scores fell below a pre-specified threshold were discarded. This simple yet effective generate–score–filter procedure enabled us to expand the dataset while preserving format compliance and overall quality.

\subsection{Instruction Tuning and Alignment}
For SFT, we adopt the standard auto-regressive next-token prediction as training objective, which is widely used in training for LLMs. For alignment training, we conducted experiments with Direct Preference Optimization~(DPO)~\cite{rafailov2023direct} and Kahneman-Tversky Optimization~(KTO)~\cite{ethayarajh2024kto} using a preference dataset constructed from a mixture of alignment datasets. Our empirical results showed that DPO consistently yields better performance and generates more appropriate responses compared to KTO, resulting that we chose DPO as our primary alignment tuning method. Moreover, we employ a two-stage training strategy to further enhance the the model's preference alignment capabilities. Each stage is trained on a distinct subset of data, and the second stage resumes training from the best checkpoint obtained in the first stage. We posited that this curriculum mitigates the tendency of DPO to overfit to preference signal, thereby facilitating more reliable model's alignment behavior.

\section{Evaluation}

\begin{table}[hbt!] 
    \centering
    \resizebox{0.8\textwidth}{!}{
    \begin{tabular}{l r r r}
        \toprule
        \makecell[l]{\textbf{Baseline} \\ \textbf{Model}} & \makecell[r]{\textbf{Parameter} \\ \textbf{Size}} &
        \makecell[r]{\textbf{Average} \\ \textbf{Improvement}} & \makecell[r]{\textbf{Average LC} \\ \textbf{Improvement}} \\
        \midrule
        Mistral & 7B & +25.47\% & +28.40\% \\
        Gemma 1 & 2B & +87.20\% & +91.77\% \\
        Gemma 1 & 7B & +2.44\% & +4.27\% \\
        Gemma 2 & 2B & +44.07\% & +47.14\% \\
        Gemma 2 & 9B & -14.17\% & -13.27\% \\
        Gemma 3 & 1B & +27.51\% & +32.96\% \\
        Gemma 3 & 4B & +15.88\% & +14.82\% \\
        Llama 3 & 8B & -18.19\% & -14.06\% \\
        Llama 3.2 & 1B & +41.82\% & 41.18\% \\
        Llama 3.2 & 3B & -2.49\% & -2.94\% \\
        Phi-2 & 2.7B & +9.72\% & +9.15\% \\
        Phi-3 & 3.8B & -9.86\% & -9.18\% \\
        Phi-3 & 7B & -13.25\% & -13.17\% \\
        \bottomrule
    \end{tabular}
    }
    \caption{Average performance improvement of Motif-2.6B and Motif-2.6B-LC over various baseline models. LC stands for long context.
    Full evaluation results are listed in Appendix~\S\ref{sec:full_eval_reulsts}}
    \label{tab:average_performance_comparison}
\end{table} 
In evaluating Motif-2.6B, we emphasized reliable and trustworthy benchmarking by replicating the specific evaluation settings used by each comparative model, despite recognizing the potential disadvantage this approach could impose on our results. Recognizing that existing models often report benchmarks with varying settings, such as different numbers of shots (n-shot) or inclusion of CoT reasoning, we carefully replicated each model’s reported evaluation settings to ensure a fair and consistent comparison. Models compared against Motif-2.6B are Phi, Gemma, Llama, and Mistral, which are other widely used open-source models. It is worth noting that we intentionally excluded \texttt{Qwen3} from direct comparison, as it is primarily a reasoning-oriented model evaluated under specialized reasoning settings. We plan to include \texttt{Qwen3} in future comparisons once we release the reasoning-focused model we are currently developing.

The average results of our benchmarks are presented in Table \ref{tab:average_performance_comparison}, with detailed and comprehensive tables available in the Appendix~\S\ref{sec:full_eval_reulsts}. Through this approach, we validated that Motif-2.6B consistently matches or surpasses the performance of similarly sized, globally leading models, confirming its effectiveness and robustness across diverse tasks. This rigorous evaluation methodology ensures meaningful and fair comparisons, highlighting that Motif-2.6B consistently meets or exceeds the performance of similarly sized state-of-the-art models across diverse benchmarks. Based on this observations, Motif-2.6B significantly advances the landscape of efficient, scalable, and powerful foundational LLMs, offering valuable insight and a robust foundation for future research and deployment.

\section{Conclusion}

In this technical report, we presented Motif-2.6B, a foundation language model specifically designed to balance performance with computational efficiency, thereby democratizing advanced LLM capabilities. Our comprehensive experiments validated the effectiveness of novel architectural elements such as Differential Attention and PolyNorm activation functions, which collectively contributed to remarkable improvements in long-context comprehension, hallucination mitigation, and in-context learning.

Our two-stage dynamic mixture pre-training strategy, complemented by the novel use of data scheduling and context length extension via RoPE adjustments, proved highly effective in enhancing the model's generalization and domain-specific reasoning abilities. Furthermore, dataset curation in the post-training phase, combining both human-annotated and synthetic datasets, along with alignment through DPO, substantially aligned the model with practical human preferences and real-world applications.

Evaluation results demonstrate that Motif-2.6B consistently meets or exceeds performance benchmarks relative to similarly sized models, achieving particularly strong results across various challenging tasks. This positions Motif-2.6B as an impactful contribution to the landscape of efficient and scalable foundational models. Looking ahead, future research will explore the extension of these methodologies to reasoning-oriented models, further optimizing architectural elements, and broadening real-world applicability.

%% file: appendix.tex
\appendix
\section{Appendix}
\label{sec:appendix}

\subsection{Contributions}

All authors are sorted alphabetically by last name.

\textbf{Technical and management leadership}: Sungmin Lee, Junghwan Lim

\textbf{Core contributors}: Dongseok Kim, Eunhwan Park, Hyunbyung Park, Junhyeok Lee

\textbf{Contributors}: Wai Ting Cheung, Dahye Choi, Jaeheui Her, Jaeyeon Huh, Hanbin Jung, Changjin Kang, Beomgyu Kim, Jihwan Kim, Minjae Kim, Taehwan Kim, Youngrok Kim, Haesol Lee, Jeesoo Lee, Kungyu Lee, Dongpin Oh, Yeongjae Park, Bokki Ryu, Daewon Suh, Dongjoo Weon

\subsection{Full Evaluation Results}
\label{sec:full_eval_reulsts}

\begin{table}[H]
    \centering
    \resizebox{1.0\textwidth}{!}{
    \begin{tabular}{l rrrrrr}
        %\toprule

        %\textbf{Benchmark} & \textbf{Metric} & \textbf{Mistral 7B} & \textbf{Motif 2.6B} &  \textbf{Motif 2.6B LC} & $\boldsymbol{\Delta}$ & $\boldsymbol{\Delta}$\textbf{LC} \\
        %\midrule

        \toprule
        \textbf{Benchmark} &
        \textbf{Metric} &
        \textbf{Mistral 7B} & 
        \multicolumn{2}{c}{\textbf{Motif}} & 
        \multicolumn{2}{c}{$\boldsymbol{\Delta}$\textbf{Mistral 7B}} \\ 
        \cmidrule(lr){4-5} \cmidrule(lr){6-7}
        & & &
        \textbf{2.6B} & \textbf{2.6B LC} & 
        \textbf{2.6B} & \textbf{2.6B LC} \\
        \midrule

        MMLU       & 5-shot                 & 60.1  & 58.0  & 58.7  & -3.61\%  &  -2.36\% \\
        HellaSwag  & 0-shot                 & 81.3  & 61.4  & 60.4  & -24.54\%   &  -25.76\% \\
        WinoG      & 0-shot                 & 75.3  & 59.9  & 60.3  & -20.44\%   &  -19.93\% \\
        PIQA       & 0-shot                 & 83.0  & 76.0  & 74.6  & -8.49\%    &  -10.13\% \\
        ARC-E      & 0-shot                 & 80.0  & 87.2  & 84.7  & +9.01\%    &  +5.81\% \\
        ARC-C      & 0-shot                 & 55.5  & 74.2   & 73.0  & +33.69\%   &  +31.44\% \\
        NQ         & 5-shot                 & 28.8  & 11.1  & 8.3  & -61.32\%   &  -71.35\% \\
        TriviaQA   & 5-shot                 & 69.9  & 55.0  & 50.0  & -21.36\%   &  -28.43\% \\
        HumanEval  & 0-shot                 & 30.5  & 68.3   & 70.1  & +123.93\%  &  +129.84\% \\
        MBPP       & 3-shot                 & 47.5  & 60.3   & 60.1  & +26.95\%   &  +26.53\% \\
        MATH       & 4-shot, maj@4$^{1}$    & 13.1  & 40.2   & 47.3  & +206.87\%  &  +260.92\% \\
        GSM8K      & 8-shot, maj@8$^{1}$    & 52.2  & 75.7  & 75.3  & +53.66\%   &  +44.21\% \\
        \midrule
        \midrule
                   &                        &       &       & \textbf{Average} & \textbf{+25.47\%} & \textbf{+28.40\%} \\
        \bottomrule
    \end{tabular}
    }
    \caption{Performance comparison between Mistral 7B and Motif 2.6B across various benchmarks. The evaluation settings and performance scores for all compared models are taken from their respective technical reports. Note that we employed maj@1 for $4$ and $8$ shot examples instead of employing maj@4 for 4 shot examples.}
    \label{tab:benchmark_comparison_mistral}
\end{table}

\begin{table}[H]
    \centering
    %\fontsize{6.8}{7}\selectfont
    %\setlength{\tabcolsep}{1.1pt}
    %\renewcommand{\arraystretch}{1.1}
    \resizebox{\linewidth}{!}{
    \begin{tabular}{@{}l r r r r r r r r r @{}}
        % \toprule
        % \textbf{Benchmark} & \textbf{Metric} & \makecell[r]{\textbf{Gemma} \\ \textbf{1 2B}} & \makecell[r]{\textbf{Gemma} \\ \textbf{1 7B}} & \makecell[r]{\textbf{Gemma} \\ \textbf{2 2B}} & \makecell[r]{\textbf{Gemma} \\ \textbf{2 9B}} & \makecell[r]{\textbf{Motif} \\ \textbf{2.6B}} & \makecell[r]{\textbf{Motif} \\ \textbf{2.6B LC}} & \makecell[r]{\textbf{Imp. over} \\ \textbf{Gemma 1} \\ \textbf{2B}} & \makecell[r]{\textbf{LC Imp.} \\ \textbf{over} \\ \textbf{Gemma 1 2B}} & \makecell[r]{\textbf{Imp. over} \\ \textbf{Gemma 1 7B}} & \makecell[r]{\textbf{LC Imp.} \\ \textbf{over} \\ \textbf{Gemma 1 7B}}  & \makecell[r]{\textbf{Imp. over} \\ \textbf{Gemma 2} \\ \textbf{2B}} & \makecell[r]{\textbf{LC Imp.} \\ \textbf{over} \\ \textbf{Gemma 2 2B}} & \makecell[r]{\textbf{Imp. over} \\ \textbf{Gemma 2 9B}} & \makecell[r]{\textbf{LC Imp.} \\ \textbf{over} \\ \textbf{Gemma 2 9B}} \\
        % \midrule

        \toprule
        \textbf{Benchmark} & 
        \textbf{Metric} & 
        \multicolumn{2}{c}{\textbf{Gemma 1}} & 
        \multicolumn{2}{c}{\textbf{Motif}} & 
        \multicolumn{2}{c}{$\boldsymbol{\Delta}$\textbf{Gemma 1 2B}} &
        \multicolumn{2}{c}{$\boldsymbol{\Delta}$\textbf{Gemma 1 7B}} \\
        \cmidrule(lr){3-4} \cmidrule(lr){5-6} \cmidrule(lr){7-8} 
        \cmidrule(lr){9-10}
        & \textbf{\#shot} & 
        \textbf{2B} & \textbf{7B} & 
        \textbf{2.6B} & \textbf{2.6B LC} & 
        \textbf{2.6B} & \textbf{2.6B LC} & 
        \textbf{2.6B} & \textbf{2.6B LC} \\
        \midrule
        
        MMLU & 5 & 42.3 & 64.4 & 58.0 & 58.7 & +36.95\% & +38.72\% & +10.98\% & +12.41\% \\
        ARC-C & 25 & 48.5 & 61.1 & 75.0 & 74.0 & +54.80\% & +52.70\% & +34.79\% & +32.96\% \\
        GSM8K & 5 & 15.1 & 51.8 & 75.13 & 74.8 & +397.55\% & +395.03\% & +209.18\% & +207.61\% \\
        AGIEval & 3-5 & 24.2 & 44.9 & 30.9 & 31.0 & +27.64\% & +28.14\% & -1.94\% & -1.56\% \\
        DROP & 3, F1 & 48.5 & 56.3 & 44.3 & 39.3 & -8.70\% & -18.93\% & -13.52\% & -23.20\%  \\
        BBH & 3, CoT & 35.2 & 59 & 48.6 & 46.4 & +37.95\% & +31.93\% & +15.89\% & +10.84\% \\
        Winogrande & 5 & 66.8 & 79 & 67.1 & 66.5 & +0.43\% & -0.52\% & -5.90\% & -6.80\% \\
        HellaSwag & 10 & 71.7 & 82.3 & 69.9 & 70.3 & -2.52\% & -1.85\% & -4.13\% & -3.47\% \\
        MATH & 4 & 11.8 & 24.3 & 40.2 & 47.3 & +240.68\% & +300.68\% & +151.25\% & +195.50\% \\
        ARC-E & 0 & 73.2 & 81.5 & 87.21 & 84.7 & +19.14\% & +15.64\% & +8.20\% & +5.02\% \\
        PIQA & 0 & 77.3 & 81.2 & 75.95 & 74.6 & -1.75\% & -3.51\% & -3.12\% & -4.86\% \\
        SIQA & 0 & 49.7 & 51.8& 61.97 & 63.3 & +24.69\% & +27.26\% & +19.40\% & +21.87\% \\
        BoolQ & 0 & 69.4 & 83.2 & 67.76 & 71.0 & -2.36\% & 2.31\% & -6.80\% & -2.34\% \\
        TriviaQA & 5 & 53.2 & 63.4 & 54.97 & 50.0 & +3.33\% & -5.96\% & -8.99\% & -17.17\% \\
        NQ & 5 & 12.5 & 23 & 11.14 & 8.25 & -12.72\% & -34.00\% & -36.20\% & -51.75\% \\
        HumanEval & P@1 & 22 & 32.3 & 68.3 & 70.1 & +210.45\% & +218.64\% & +239.80\% & +248.76\% \\
        MBPP & 3 & 29.2 & 44.4 & 60.3& 60.1 & +106.51\% & +105.82\% & +99.67\% & +99.01\% \\
        \midrule
         &   &   &   &   & \textbf{Average} & \textbf{+87.20\%} & \textbf{+91.77\%}& \textbf{+2.54\%} & \textbf{+4.27\%} \\
        \bottomrule
    \end{tabular}
    }
    \caption{Performance comparison across Gemma 1 models and Motif 2.6B on various benchmarks. The evaluation settings and performance scores for all compared models are taken from their respective technical reports.}
    \label{tab:benchmark_comparison_gemma_1}
\end{table}

\begin{table}[H]
    \centering
    %\fontsize{6.8}{7}\selectfont
    %\setlength{\tabcolsep}{1.1pt}
    %\renewcommand{\arraystretch}{1.1}
    \resizebox{\linewidth}{!}{
    \begin{tabular}{@{}l r r r r r r r r r @{}}
        % \toprule
        % \textbf{Benchmark} & \textbf{Metric} & \makecell[r]{\textbf{Gemma} \\ \textbf{1 2B}} & \makecell[r]{\textbf{Gemma} \\ \textbf{1 7B}} & \makecell[r]{\textbf{Gemma} \\ \textbf{2 2B}} & \makecell[r]{\textbf{Gemma} \\ \textbf{2 9B}} & \makecell[r]{\textbf{Motif} \\ \textbf{2.6B}} & \makecell[r]{\textbf{Motif} \\ \textbf{2.6B LC}} & \makecell[r]{\textbf{Imp. over} \\ \textbf{Gemma 1} \\ \textbf{2B}} & \makecell[r]{\textbf{LC Imp.} \\ \textbf{over} \\ \textbf{Gemma 1 2B}} & \makecell[r]{\textbf{Imp. over} \\ \textbf{Gemma 1 7B}} & \makecell[r]{\textbf{LC Imp.} \\ \textbf{over} \\ \textbf{Gemma 1 7B}}  & \makecell[r]{\textbf{Imp. over} \\ \textbf{Gemma 2} \\ \textbf{2B}} & \makecell[r]{\textbf{LC Imp.} \\ \textbf{over} \\ \textbf{Gemma 2 2B}} & \makecell[r]{\textbf{Imp. over} \\ \textbf{Gemma 2 9B}} & \makecell[r]{\textbf{LC Imp.} \\ \textbf{over} \\ \textbf{Gemma 2 9B}} \\
        % \midrule

        \toprule
        \textbf{Benchmark} & 
        \textbf{Metric} & 
        \multicolumn{2}{c}{\textbf{Gemma 2}} & 
        \multicolumn{2}{c}{\textbf{Motif}} & 
        \multicolumn{2}{c}{$\boldsymbol{\Delta}$\textbf{Gemma 2 2B}} &
        \multicolumn{2}{c}{$\boldsymbol{\Delta}$\textbf{Gemma 2 7B}} \\
        \cmidrule(lr){3-4} \cmidrule(lr){5-6} \cmidrule(lr){7-8} 
        \cmidrule(lr){9-10}
        & \textbf{\#shot} & 
        \textbf{2B} & \textbf{7B} & 
        \textbf{2.6B} & \textbf{2.6B LC} & 
        \textbf{2.6B} & \textbf{2.6B LC} & 
        \textbf{2.6B} & \textbf{2.6B LC} \\
        \midrule
        
        MMLU & 5 & 52.2 & 71.3 & 58.0 & 58.7 & -10.05\% & -8.88\% & -18.75\% & -17.70\%  \\
        ARC-C & 25 & 55.7 & 68.4 & 75.0 & 74.0 & +22.88\% & +21.21\% & +9.77\% & +8.27\%  \\
        GSM8K & 5 & 24.3 & 68.6 & 75.13 & 74.8 & +45.04\% & +44.31\% & +9.52\% & +8.97\%  \\
        AGIEval & 3-5 & 31.5 & 52.8 & 30.9 & 31.0 & -31.20\% & -30.94\% & -41.50\% & -41.27\%  \\
        DROP & 3, F1 & 51.2 & 69.4 & 44.3 & 39.3 & -21.35\% & -30.16\% & -36.20\% & -43.34\%  \\
        BBH & 3, CoT & 41.9 & 68.2 & 48.6 & 46.4 & -17.69\% & -21.29\% & -28.80\% & -31.91\% \\
        Winogrande & 5 & 71.3 & 80.6 & 67.1 & 66.5 & -15.08\% & -15.89\% & -16.76\% & -17.56\%  \\
        HellaSwag & 10 & 72.9 & 81.9 & 69.9 & 70.3 & -15.08\% & -14.50\% & -14.66\% & -14.08\%  \\
        MATH & 4 & 16 & 36.6 & 40.2 & 47.3 & +65.43\% & +94.57\% & +9.84\% & +29.18\%  \\
        ARC-E & 0 & 80.6 & 88 & 87.21 & 84.7 & +7.01\% & +3.87\% & -0.90\% & -3.81\%  \\
        PIQA & 0 & 78.4 & 81.7 & 75.95 & 74.6 & -6.47\% & -8.14\% & -7.04\% & -8.70\%  \\
        SIQA & 0 & 51.9 & 53.4 & 61.97 & 63.3 & +19.63\% & +22.10\% & +16.05\% & +18.45\%  \\
        BoolQ & 0 & 72.7 & 84.2 & 67.76 & 71.0 & -18.56\% & -14.66\% & -19.52\% & -15.68\%  \\
        TriviaQA & 5 & 60.4 & 76.6 & 54.97 & 50.0 & -13.30\% & -21.09\% & -28.24\% & -34.69\%  \\
        NQ & 5 & 17.1 & 29.2 & 11.14 & 8.25 & -52.57\% & -64.13\% & -62.64\% & -71.75\%  \\
        HumanEval & P@1 & 20.1 & 40.2 & 68.3 & 70.1 & +111.46\% & +117.03\% & +69.90\% & +74.38\%  \\
        MBPP & 3 & 30.2 & 52.4 & 60.3& 60.1 & +35.81\% & +35.36\% & +15.08\% & +14.69\%  \\
        \midrule
          &   &   &   &   & \textbf{Average} & \textbf{+44.07\%} &  \textbf{+47.14\%} & \textbf{-14.17\%} & \textbf{ -13.27\%}\\
        \bottomrule
    \end{tabular}
    }
    \caption{Performance comparison across Gemma 2 models and Motif 2.6B on various benchmarks. The evaluation settings and performance scores for all compared models are taken from their respective technical reports.}
    \label{tab:benchmark_comparison_gemma_2}
\end{table}

\begin{table}[H]
    \centering
    % \caption{Performance comparison across Gemma 3 models and Motif 2.6B on various benchmarks. The evaluation settings and performance scores for all compared models are taken from their respective technical reports.}
    \resizebox{\linewidth}{!}{
    \begin{tabular}{@{}l r r r r r r r r r@{}}
        % \toprule
        % \textbf{Benchmark} & \textbf{Metric} & \makecell[r]{\textbf{Gemma 3} \\ \textbf{1B}} & \makecell[r]{\textbf{Gemma 3} \\ \textbf{4B}} & \textbf{Motif 2.6B} & \makecell[r]{\textbf{Motif} \\ \textbf{2.6B LC}} & \makecell[r]{\textbf{Imp. over} \\ \textbf{Gemma 3 1B}} & \makecell[r]{\textbf{LC Imp.} \\  \textbf{over} \\ \textbf{Gemma 3 1B}}  & \makecell[r]{\textbf{Imp over} \\ \textbf{Gemma 3 4B}}  & \makecell[r]{\textbf{LC Imp.} \\  \textbf{over} \\ \textbf{Gemma 3 4B}} \\
        % \midrule

        \toprule
        \textbf{Benchmark} & 
        \textbf{Metric} & 
        \multicolumn{2}{c}{\textbf{Gemma 3}} & 
        \multicolumn{2}{c}{\textbf{Motif}} & 
        \multicolumn{2}{c}{$\boldsymbol{\Delta}$\textbf{Gemma 3 1B}} & 
        \multicolumn{2}{c}{$\boldsymbol{\Delta}$\textbf{Gemma 3 4B}} \\
        \cmidrule(lr){3-4} \cmidrule(lr){5-6} \cmidrule(lr){7-8} \cmidrule(lr){9-10}
        & & 
        \textbf{1B} & \textbf{4B} & 
        \textbf{2.6B} & \textbf{2.6B LC} & 
        \textbf{2.6B} & \textbf{2.6B LC} & 
        \textbf{2.6B} & \textbf{2.6B LC} \\
        \midrule

        HellaSwag & 10-shot & 62.3 & 77.2 & 69.9 & 70.3 & +12.18\% & +12.90\% & -9.47\% & -9.46\% \\
        BoolQ & 0-shot & 63.2 & 72.3 & 67.8 & 71.0 & +7.22\% & +12.34\% & -6.28\% & -1.80\%\\
        PIQA & 0-shot & 73.8 & 79.6 & 75.6 & 74.6 &+2.43\% & +1.08\% & -5.04\% & -6.28\% \\
        SIQA & 0-shot & 48.9 & 51.9 & 62.0 & 63.3 & +26.73\% & +29.45\% & +19.40\% & +21.97\% \\
        TriviaQA & 5-shot & 39.8 & 65.8 & 55.0 & 50.0 & +38.12\% & +25.63\% & -16.46\% & -24.01\%\\
        NQ & 5-shot & 9.48 & 20 & 11.1 & 8.25 & +15.08\% & -12.63\% & -45.5\% & -58.5\% \\
        ARC-C & 25-shot & 38.4 & 56.2 & 75.1 & 74.0 & +95.52\% & +92.71\% & +33.59\% & +31.67\% \\
        ARC-E & 0-shot & 73 & 82.4 & 87.2 & 84.7 &+19.47\% & +16.03\% & +5.84\% & +2.79\%\\
        Winogrande & 5-shot & 58.2 & 64.7 & 67.1 & 66.5 & +15.27\% & +14.26\% & +3.69\% & +2.78\%\\
        BBH & few-shot, CoT & 28.4 & 50.9 & 48.6 & 46.4 & +70.99\% & +63.38\% & -4.60\% & -8.84\% \\
        Drop & 1-shot, F1 & 42.4 & 60.1 & 44.3 & 39.3 & -30.83\% & -7.31\% & -51.20\% & -34.61\% \\
        MMLU & 5-shot & - & 59.6 & 58.0 & 58.7 & - & - & -2.80\% & -1.51\% \\
        MMLU-Pro & 5-shot, CoT & 14.7 & 29.2 & 28.4 & 32.0 & +93.2\% & +117.69\% & -2.74\% & +9.59\%\\
        AGIE & 3-5-shot & - & 42.1 & 30.9 & 31.0 & -  & - & -26.60\% & -26.37\% \\
        MATH & 4-shot, CoT & - & 24.2 & 40.2 & 47.3 & - & - & +66.12\% & +95.45\% \\
        GSM8K & 8-shot, CoT & - & 38.4 & 80.2 & 80.3 & - & - & +108.88\% & +109.11\% \\
        GPQA Diamond & 5-shot, CoT & - & 15 & 31.8 & 28.3 & - & - & +112.07\% & +88.67\% \\
        MBPP & 3-shot & - & 46 & 60.3 & 60.1 & - & - & +31.09\% & +30.65\% \\
        HumanE & 0-shot & - & 36 & 68.3 & 70.1 & - & - & +89.72\% & +94.72\%\\
        IFEval & 0-shot & 80.2 & 90.2 & 74.02 & 72.5 & -7.71\% & -9.60\% & -17.94\% & -19.62\% \\
        \midrule
          &   &   &   &   & \textbf{Average} & \textbf{+27.51\%} & \textbf{+27.37\%} & \textbf{+15.88\%} & \textbf{+14.82\%} \\
        \bottomrule
    \end{tabular}
    }
    \caption{Performance comparison across Gemma 3 models and Motif 2.6B on various benchmarks. The evaluation settings and performance scores for all compared models are taken from their respective technical reports. Missing values are not officially reported.
}
    \label{tab:benchmark_comparison_gemma3}
\end{table}
% \end{landscape}

\begin{table}[H]
    \centering
    % \caption{Performance comparison between Llama 3 8B and Motif 2.6B across various benchmarks. The evaluation settings and performance scores for all compared models are taken from their respective technical reports.}
    \renewcommand{\arraystretch}{1.1}
    \resizebox{\textwidth}{!}{
    \begin{tabular}{@{}l r r r r r r@{}}
        %\toprule
        %\textbf{Benchmark} &
        %\textbf{Metric} &
        %\textbf{Llama 3 8B} & 
        %\textbf{Motif 2.6B} & 
        %\makecell[r]{\textbf{Motif} \\ \textbf{2.6B LC}} &
        %$\boldsymbol{\Delta}$ &
        %$\boldsymbol{\Delta}$\textbf{LC} \\
        %\midrule

        \toprule
        \textbf{Benchmark} &
        \textbf{Metric} &
        \textbf{Llama 3 8B} & 
        \multicolumn{2}{c}{\textbf{Motif}} & 
        \multicolumn{2}{c}{$\boldsymbol{\Delta}$\textbf{Llama 3 8B}} \\ 
        \cmidrule(lr){4-5} \cmidrule(lr){6-7}
        & & &
        \textbf{2.6B} & \textbf{2.6B LC} & 
        \textbf{2.6B} & \textbf{2.6B LC} \\
        \midrule
        MMLU & 5-shot & 69.4 & 57.93 & 58.68 & -16.53\% & -15.45\% \\
        MMLU & 0-shot, CoT & 73 & 57.95 & 57.46 & -20.62\%  & -21.29\% \\
        MMLU-Pro & 5-shot, CoT & 48.3 & 28.4 & 31.94 & -41.20\% & -33.87\% \\
        IFEval & - & 80.4 & 74.02 & 72.53 & -7.94\%  & -9.79\% \\
        HumanEval & 0-shot & 72.6 & 68.3 & 70.1 & -5.92\%  & -3.44\% \\
        MBPP & 0-shot & 72.8 & 57.93 & 60.1 & -20.43\%  & -17.45\% \\
        GSM8K & 8-shot, CoT & 84.5 & 80.21 & 80.28 & -5.08\%  & -4.99\% \\
        MATH & 0-shot, CoT & 51.9 & 49.68 & 49.24 & -4.28\%  & -5.13\% \\
        ARC Challenge & 0-shot & 83.4 & 74.2 & 73.2 & -11.03\%  & -12.23\% \\
        GPQA & 0-shot, CoT & 32.8 & 18.53 & 27.23 & -43.51\%  & -16.98\% \\
        \midrule
          &   &   &   & \textbf{Average} & \textbf{-18.19\%} & \textbf{-14.06\%} \\
        \bottomrule
    \end{tabular}
    }
    \caption{Performance comparison between Llama 3 8B and Motif 2.6B across various benchmarks. The evaluation settings and performance scores for all compared models are taken from their respective technical reports.}
    \label{tab:benchmark_comparison_llama3}
\end{table}

% \begin{landscape}
\begin{table}[H]
    \centering
    % \caption{Performance comparison between Llama 3.2 models and Motif 2.6B across various benchmarks. The evaluation settings and performance scores for all compared models are taken from their respective technical reports.}
    \renewcommand{\arraystretch}{1.1}
    \resizebox{\linewidth}{!}{
    \begin{tabular}{@{}l r r r r r r r r r@{}}
        % \toprule
        % \textbf{Benchmark} & \textbf{Metric} & \textbf{Llama 3.2 1B} & \textbf{Llama 3.2 3B} & \textbf{Motif 2.6B} & \makecell[r]{\textbf{Motif} \\ \textbf{2.6B LC}} & \makecell[r]{\textbf{Imp. over} \\ \textbf{Llama 3.2 1B}} & \makecell[r]{\textbf{LC Imp. over} \\ \textbf{Llama 3.2 1B}} & \makecell[r]{\textbf{Imp. over} \\ \textbf{Llama 3.2 3B}} & \makecell[r]{\textbf{LC Imp. over} \\ \textbf{Llama 3.2 3B}} \\
        % \midrule

        \toprule
        \textbf{Benchmark} & 
        \textbf{Metric} & 
        \multicolumn{2}{c}{\textbf{Llama 3.2}} & 
        \multicolumn{2}{c}{\textbf{Motif}} & 
        \multicolumn{2}{c}{$\boldsymbol{\Delta}$\textbf{Llama 3.2 1B}} & 
        \multicolumn{2}{c}{$\boldsymbol{\Delta}$\textbf{Llama 3.2 3B}} \\
        \cmidrule(lr){3-4} \cmidrule(lr){5-6} \cmidrule(lr){7-8} \cmidrule(lr){9-10}
        & & 
        \textbf{1B} & \textbf{3B} & 
        \textbf{2.6B} & \textbf{2.6B LC} & 
        \textbf{2.6B} & \textbf{2.6B LC} & 
        \textbf{2.6B} & \textbf{2.6B LC} \\
        \midrule
                
        MMLU & 0-shot & 49.3 & 63.4 & 57.56 & 58.68 & +16.75\% & +19.03\% & -9.21\% & -7.44\% \\
        IFEval & - & 59.5 & 77.4 & 74.02 & 72.53 & +24.40\% & +21.89\% & -4.37\%  & -6.3\%\\
        GSM8K & 8-shot, CoT & 44.4 & 77.7 & 80.21 & 80.28 & +80.65\% & +80.81\% & +3.23\%  & +3.32\%\\
        MATH & 0-shot, CoT & 30.6 & 48 & 49.68 & 49.24 & +62.35\% & +60.92\% & +3.50\%  & +2.58\%\\
        ARC Challenge & 0-shot & 59.4 & 78.6 & 74.2 & 73.2 & +24.92\% & +23.23\% & -5.60\%  & -6.87\%\\
        GPQA & 0-shot & 27.2 & 32.8 & 25.45 & 27.23 & -6.43\% & +0.11\% & -22.41\%  & -16.98\% \\
        Hellaswag & 0-shot & 41.2 & 69.8 & 61.35 & 60.36 & +48.91\% & +46.5\% & -12.11\%  & -13.52\%\\
        \midrule
          &   &   &   &   & \textbf{Average} & \textbf{+41.82\%} & \textbf{+41.18\%} & \textbf{-2.49\%} & \textbf{-2.94\%} \\
        \bottomrule
    \end{tabular}
    }
    \caption{Performance comparison between Llama 3.2 models and Motif 2.6B across various benchmarks. The evaluation settings and performance scores for all compared models are taken from their respective technical reports.}
    \label{tab:benchmark_comparison_llama32}
\end{table}
\begin{table}[H]
    \centering
    % \caption{Performance comparison between Phi-3 and Phi-2 models and Motif 2.6B across various benchmarks. The evaluation settings and performance scores for all compared models are taken from their respective technical reports.}
    \renewcommand{\arraystretch}{1.1}
    \resizebox{\linewidth}{!}{
    \begin{tabular}{@{}l r r r r r r r r r r r r@{}}
        % \toprule
        % \textbf{Benchmark} & \textbf{Metric} & \makecell[r]{\textbf{Phi-2} \\ \textbf{2.7B}} & \makecell[r]{\textbf{Phi-3} \\ \textbf{3.8B}} & \makecell[r]{\textbf{Phi-3} \\ \textbf{7B}} & \makecell[r]{\textbf{Motif} \\ \textbf{2.6B}} & \makecell[r]{\textbf{Motif} \\ \textbf{2.6B LC}} &  \makecell[r]{\textbf{Imp.} \\ \textbf{over} \\ \textbf{Phi-2 2.7B}} & \makecell[r]{\textbf{LC Imp.} \\ \textbf{over} \\ \textbf{Phi-2 2.7B}} & \makecell[r]{\textbf{Imp.} \\ \textbf{over} \\ \textbf{Phi-3 3.8B}} & \makecell[r]{\textbf{LC Imp.} \\ \textbf{over} \\ \textbf{Phi-3 3.8B}} & \makecell[r]{\textbf{Imp.} \\ \textbf{over} \\ \textbf{Phi-3 7B}}  & \makecell[r]{\textbf{LC Imp.} \\ \textbf{over} \\ \textbf{Phi-3 7B}} \\
        % \midrule

        \toprule
        \textbf{Benchmark} & 
        \textbf{Metric} & 
        \multicolumn{3}{c}{\textbf{Phi}} & 
        \multicolumn{2}{c}{\textbf{Motif}} & 
        \multicolumn{2}{c}{$\boldsymbol{\Delta}$\textbf{Phi-2 2.7B}} & 
        \multicolumn{2}{c}{$\boldsymbol{\Delta}$\textbf{Phi-3 3.8B}} & 
        \multicolumn{2}{c}{$\boldsymbol{\Delta}$\textbf{Phi-3 7B}} \\
        \cmidrule(lr){3-5} \cmidrule(lr){6-7} \cmidrule(lr){8-9} \cmidrule(lr){10-11} \cmidrule(lr){12-13}
        & & 
        \textbf{2.7B} & \textbf{3.8B} & \textbf{7B} & 
        \textbf{2.6B} & \textbf{2.6B LC} & 
        \textbf{2.6B} & \textbf{2.6B LC} & 
        \textbf{2.6B} & \textbf{2.6B LC} & 
        \textbf{2.6B} & \textbf{2.6B LC} \\
        \midrule
        
        MMLU & 5-shot & 56.3 & 68.8 & 75.7 & 57.93 & 58.68 & +2.90\% & +4.23\% & -15.80\% & -14.71\% & -23.47\% & -22.48\% \\
        HellaSwag & 5-shot & 53.6 & 76.7 & 77 & 68.97 & 69.39 & +28.68\% & +29.46\%  &-10.08\% & -9.53\% & -10.43\% & -9.88\% \\
        ANLI & 7-shot & 42.5 & 52.8 & 58.1 & 47.99 & 49.1 & +12.92\% & +15.53\%  & -9.11\% & -7.01\% & -17.40\% & -15.49\% \\
        GSM-8K & 8-shot, CoT & 61.1 & 82.5 & 89.6 & 75.66 & 80.21 & +23.83\%  & +31.28\% & -8.29\% & -2.78\% & -15.56\% & -10.48\% \\
        MATH & 0-shot, CoT & - & 41.3 & 34.6 & 49.68 & 49.24 & -  & - & +20.29\% & +19.23\% & +43.58\% & +42.31\% \\
        MedQA & 2-shot & 40.9 & 53.8 & 65.4 & 42.1 & 42.02 & +2.93\%  & +2.74\% & -21.75\% & -21.9\% & -35.63\% & -35.75\% \\
        AGIEval & 0-shot & 29.8 & 37.5 & 45.1 & 30.89 & 31.01 & +3.66\%  & +4.06\% & -17.63\% & -17.31\% & -31.51\% & -31.24\% \\
        TriviaQA & 5-shot & 45.2 & 64 & 58.1 & 54.97 & 50.03 & +21.62\%  & +10.69\% & -14.11\% & -21.83\% & -5.39\% & -13.89\% \\
        Arc-C & 10-shot & 75.9 & 84.9 & 90.7 & 75.17 & 75.34 & -0.96\%  & -0.74\% & -11.46\% & -11.26\% & -17.12\% & -16.93\% \\
        Arc-E & 10-shot & 88.5 & 94.6 & 97 & 88.64 & 87.08 & +0.16\% & -1.6\% & -6.30\% & -7.95\% & -8.62\% & -10.23\% \\
        PIQA & 5-shot & 60.2 & 84.2 & 86.9 & 78.29 & 77.09 & +30.05\%  & +28.06\% & -7.02\% & -8.44\% & -9.91\% & -11.29\% \\
        SociQA & 5-shot & 68.3 & 76.6 & 79.2 & 66.73 & 65.35 & -2.30\%  & -4.32\% & -12.89\% & -14.69\% & -15.74\% & -17.49\% \\
        BigBench-Hard & 3-shot, CoT & 59.4 & 71.7 & 79.1 & 48.56 & 46.44 & -18.25\%  & -21.82\% & -32.27\% & -35.23\% & -38.61\% & -41.29\% \\
        WinoGrande & 5-shot & 54.7 & 70.8 & 81.5 & 67.09 & 66.45 & +22.65\%  & +21.48\% & -5.24\% & -6.14\% & -17.68\% & -18.47\% \\
        OpenBookQA & 10-shot & 73.6 & 83.2 & 88 & 87.8 & 87.2 & +19.29\%  & +18.48\% & +5.53\% & +4.81\% & -0.23\% & -0.91\% \\
        BoolQ & 2-shot & - & 77.2 & 84.8 & 70.7 & 74.43 & -  & - & -8.42\% & -3.59\% & -16.63\% & -12.23\% \\
        CommonSenseQA & 10-shot & 69.3 & 80.2 & 80 & 71.25 & 69.36 & +2.81\%  & +0.09\% & -11.16\% & -13.52\% & -10.94\% & -13.3\% \\
        TruthfulQA & 10-shot & - & 65 & 70.2 & 52.07 & 53.49 & -  & - & -19.89\% & -17.71\% & -25.83\% & -23.8\% \\
        HumanEval & 0-shot & 59 & 58.5 & 61 & 68.29 & 70.1 & +15.75\%  & +18.81\% & +16.74\% & +19.83\% & +11.95\% & +14.92\% \\
        MBPP & 3-shot & 60.6 & 70 & 71.7 & 60.3 & 60.1 & -0.50\%  & -0.83\% & -13.86\% & -14.14\% & -15.90\% & -16.18\% \\
        GPQA & 2-shot, CoT & - & 32.8 & 34.3 & 27.9 & 32.36 & -  & - & -14.94\% & -1.34\% & -18.66\% & -5.66\% \\
        MT Bench & 2R. Avg. & - & 8.38 & 8.7 & 6.77 & 6.97 & -  & - & -19.21\% & -16.83\% & -22.18\% & -19.89\% \\
        \midrule
          &   &   &   &   &   & \textbf{Average} & \textbf{+9.72\%} & \textbf{+9.15\%} & \textbf{-9.86\%} & \textbf{-9.18\%} & \textbf{-13.72\%} & \textbf{-13.17\%} \\
        \bottomrule
    \end{tabular}
    }
    \caption{Performance comparison between Phi-3 and Phi-2 models and Motif 2.6B across various benchmarks. The evaluation settings and performance scores for all compared models are taken from their respective technical reports. Missing values are not officially reported.}
    \label{tab:benchmark_comparison_phi}
\end{table}
% \end{landscape}

% \end{comment}